\documentclass[runningheads]{llncs}

\usepackage{eccv}

\usepackage{eccvabbrv}

\usepackage{graphicx}
\usepackage{booktabs}

\usepackage[accsupp]{axessibility}  

\usepackage{multirow}
\usepackage[table]{xcolor}

\usepackage[breaklinks,colorlinks,citecolor=eccvblue]{hyperref}

\usepackage{orcidlink}

\begin{document}

\title{Representation-driven Endoscopic Visual Embedding Alignment for Latent Generation} 

\titlerunning{REVEAL}

\author{Francisco Caetano\inst{1}\orcidlink{0000-0002-6069-6084} \and Tim J.M. Jaspers\inst{1}\orcidlink{0009-0001-8306-5058} \and Haiko Middeljans\inst{1} \and Martijn R. Jong\inst{2}\orcidlink{0000-0002-8593-3875} \and Rixta A.H. van Eijck van Heslinga\inst{2} \and Floor Slooter\inst{2} \and Albert J. de Groof~\inst{2} \and Jacques J. Bergman\inst{2} \and Peter H.N. De~With\inst{1} \and Fons van der Sommen\inst{1}\orcidlink{0000-0002-3593-2356}}

\authorrunning{F.~Caetano et al.}

\institute{Department of Electrical Engineering, ARIA Lab, Eindhoven University of Technology, Eindhoven, The Netherlands\newline \url{{f.caetano}@tue.nl} \and Department of Gastroenterology and Hepatology, Amsterdam University Medical Centers, University of Amsterdam, Amsterdam, The Netherlands}

\maketitle

\begin{abstract}
Developing foundation generative models for endoscopy is limited by the gap between natural and clinical images and the computational cost of training large Diffusion Transformers. Although representation alignment has improved efficiency in general computer vision, its role within the highly specialized endoscopic image space remains unclear. We introduce REVEAL~(Representation-driven Endoscopic Visual Embedding Alignment), the largest generative foundation model for endoscopy to date, trained on GastroNet-5M~(GN-5M), a multicenter dataset of 5 million endoscopic frames. Instead of depending on out-of-domain priors, REVEAL employs encoders pretrained directly on the endoscopic distribution to align diffusion latents with domain-specific visual features, preserving fine textures and intricate anatomical structures. Beyond image generation, REVEAL also serves as a powerful feature extractor; in multiple benchmarks, it delivers performance that is competitive with, and in several cases exceeds, endoscopic foundation models such as EndoViT and Endo-FM, specifically tuned for classification tasks, while demonstrating strong representation robustness under realistic imaging corruptions. REVEAL produces high-fidelity images and maintains robust structural coherence in latent-space edits such as inpainting and outpainting. This high-capacity backbone lowers the computational threshold for building specialized clinical tools, offering an open, versatile foundation for conditional synthesis, segmentation, and out-of-distribution detection in future intelligent gastroenterology systems. The code and weights are available at \href{https://caetas.github.io/reveal.html}{caetas.github.io/reveal.html}.
\end{abstract}

\section{Introduction}
\label{sec:intro}

High-fidelity synthetic image generation offers a direct route to addressing persistent challenges in endoscopy, including limited data diversity and bias, patient privacy constraints on data sharing, and severe class imbalance for rare pathologies. Representation alignment has emerged as a crucial technique for enhancing the training efficiency and convergence of Diffusion Transformers~(DiTs)~\cite{ma2024sit}. By aligning internal latent representations with pretrained self-supervised visual encoders, recent generative frameworks have achieved significant gains in both sampling quality and data efficiency~\cite{yurepresentation}. Recent work has identified that these improvements are driven not only by global semantic information, often proxied by ImageNet-1K~\cite{deng2009imagenet} performance, but also by the preservation of complex spatial structures~\cite{singh2025matters,leng2025repa} and token-level relationships during the generative process.

This distinction is vital in the context of endoscopic imaging. Endoscopic scenes are characterized by subtle mucosal textures, specular reflections, and intricate anatomical morphologies, in which global semantic labels provide insufficient signal. Prevailing alignment strategies that rely on general-purpose vision encoders assume that semantic richness is the primary driver of synthesis quality. We argue that for medical foundation models, this reliance on out-of-domain semantic proxies may lead to suboptimal morphological fidelity and a failure to capture the clinical nuances of the endoscopic manifold.

In this paper, we introduce REVEAL, \textbf{R}epresentation-driven \textbf{E}ndoscopic \textbf{V}isual \textbf{E}mbedding \textbf{Al}ignment, a foundation generative model for endoscopy. To ground the model in domain-specific features, we leverage GastroNet-5M~\cite{jong2025gastronet}, a curated dataset of 5~million endoscopic frames. REVEAL utilizes encoders pretrained directly on this large-scale endoscopic distribution, ensuring that the target representations are inherently aligned with the unique visual characteristics of endoscopic data~\cite{boers2024foundation}. By performing representation alignment on this distribution, REVEAL learns to synthesize high-fidelity gastrointestinal landscapes that adhere to the spatial constraints of the endoscopic environment.

This paper provides the following specific contributions: (a)~an extensive benchmark of representation alignment design elements, evaluating the influence of target representations and alignment strategies on the morphological fidelity and convergence of DiTs; (b)~an evaluation of model features across multiple independent endoscopic datasets, including a corrupted robustness benchmark with realistic imaging artifacts, demonstrating generalization beyond the training distribution while maintaining anatomical integrity; and (c)~the introduction of the largest foundation generative model for endoscopy to date, a high-capacity backbone trained on GN-5M that simultaneously serves as a competitive feature extractor, exceeding established endoscopic foundation models such as EndoViT~\cite{batic2024endovit} and Endo-FM~\cite{wang2023foundation}, and enables immediate application to downstream tasks such as inpainting, outpainting, and specialized conditional generation.

\section{Related Work}

\subsection{Endoscopy Image Synthesis}

Current endoscopic image synthesis remains heavily reliant on conventional architectures such as Generative Adversarial Networks~(GANs) and Variational Autoencoders~(VAEs), primarily for data augmentation and overcoming clinical label scarcity. Recent frameworks like MSVQ-VAE~\cite{diamantis2025multiscale} and TIDE~\cite{diamantis2024intestine} continue to utilize these methods to simulate capsule endoscopy and intestinal environments; however, these models frequently struggle with training instability and preserving high-frequency mucosal textures~\cite{chu2020smoothness,zhangdimensional}. While the field has recently transitioned toward more advanced diffusion-based frameworks, a prominent line of work has focused narrowly on polyp image synthesis for segmentation and detection augmentation~\cite{machavcek2023mask,dorjsembe2024polyp,liu2025polyp}, addressing a single lesion type rather than the broader endoscopic manifold. Beyond polyp-specific generation, significant diffusion-based efforts are largely confined to the temporal domain for surgical video simulation~\cite{li2024endora} and depend on out-of-domain priors from models like Stable Diffusion~\cite{kaleta2024minimal,sharma2024controlpolypnet} or small-scale datasets for training~\cite{jaspers2025robust}. Since these models are typically pretrained on natural imagery, they lack an inherent understanding of the endoscopic manifold, resulting in a persistent domain gap that necessitates the development of generative backbones trained directly on large-scale clinical distributions.

\subsection{Endoscopy Foundation Models}
While foundation models perform strongly across a wide range of computer vision tasks~\cite{oquab2023dinov2,simeoni2025dinov3}, domain mismatch limits their effectiveness in medical applications~\cite{boers2024foundation}. Consequently, several endoscopy-specific foundation models have been developed to better capture the unique, domain-specific characteristics of endoscopic data. Notable endoscopy foundation models such as Endo-FM~\cite{wang2023foundation} and EndoViT~\cite{batic2024endovit}, which adapt masked image modeling pretraining to endoscopic data, demonstrate clear advantages over general-purpose encoders on downstream clinical tasks. More recently, EndoMamba~\cite{tian2025endomamba} proposed a Mamba-based backbone with hierarchical self-supervised pretraining on endoscopic video sequences, achieving strong performance across spatiotemporal tasks such as surgical phase recognition and localization. EndoFM-LV~\cite{wang2025improving} similarly targets long-sequence video pretraining, extending Endo-FM to minute-level clips. GN-5M~\cite{jong2025gastronet} has further enabled the development of models that surpass general-purpose foundation models with substantially less compute. Despite these advances, limited diversity in medical data continues to be a major bottleneck, constraining the development of more robust and generalizable models. Critically, all existing endoscopy foundation models are purely discriminative; no prior work has explored large-scale generative pretraining as a path toward clinical visual representations, which is the gap REVEAL addresses.

\subsection{Representation Alignment for Generative Models}

Representation alignment~(REPA) has emerged as a powerful paradigm for accelerating the training of diffusion and flow-matching transformers. REPA~\cite{yurepresentation} pioneered this direction by aligning the noisy hidden states of a denoising network with clean-image features from a frozen self-supervised encoder (\eg, DINOv2),
yielding over $17\times$ training speedup on ImageNet generation. Subsequent work has extended this idea in several directions: REPA-E~\cite{leng2025repa} unlocks end-to-end joint training of the VAE and diffusion model via the alignment loss, while VA-VAE~\cite{yao2025reconstruction} applies vision-foundation-model alignment directly to the VAE latent space. Recently, iREPA~\cite{singh2025matters} further showed that spatial structure, rather than global semantic quality, is the key property of the teacher representation that drives generation performance. Our generative model builds on iREPA, extending representation alignment to the endoscopy domain with domain-adapted vision encoders.

\section{REVEAL}

\begin{figure}[!t]
    \centering
    \includegraphics[width=\linewidth]{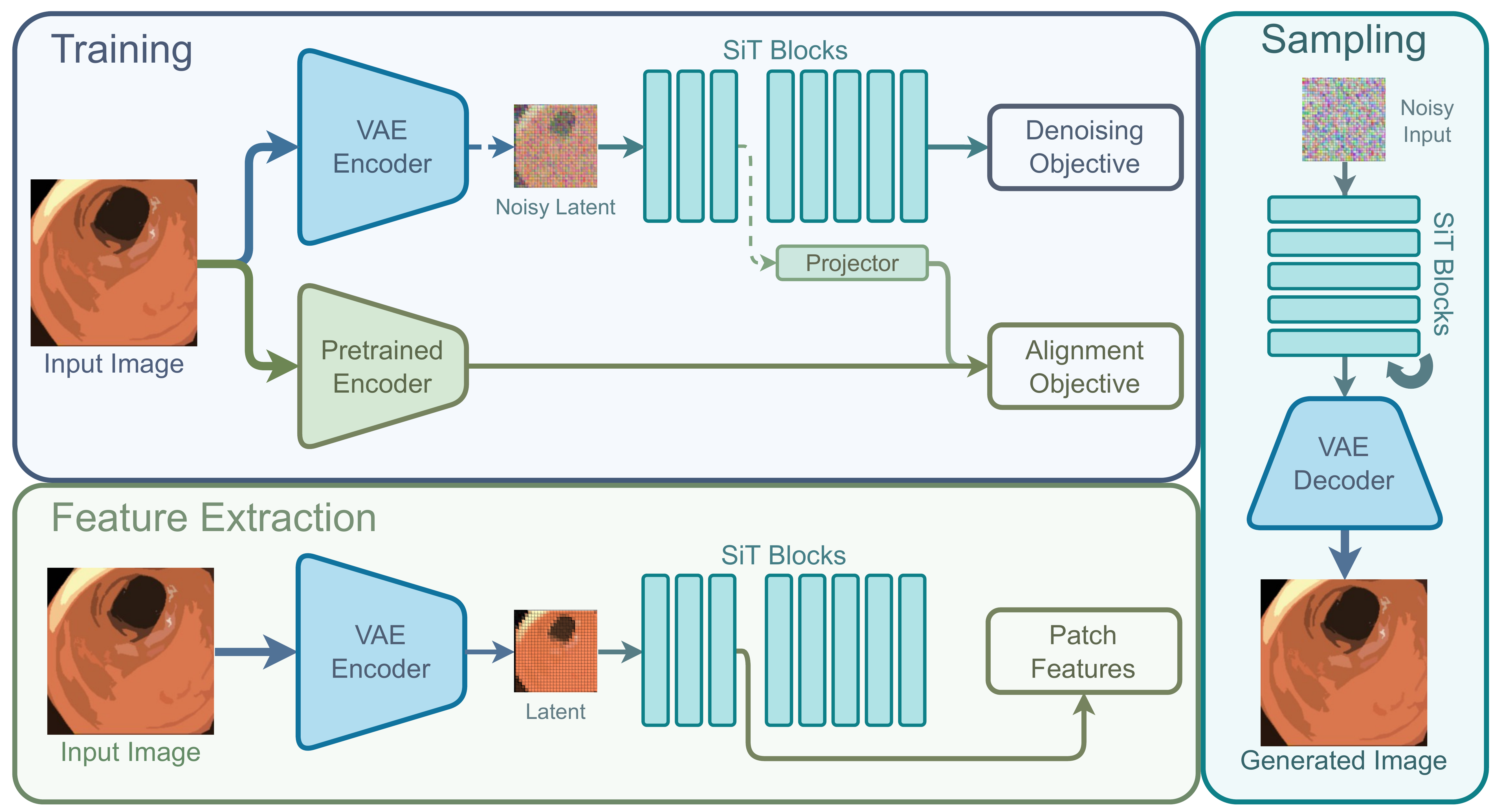}
    \caption{The REVEAL framework. Training~(top) optimizes SiT blocks using joint denoising and feature alignment objectives. Feature extraction~(bottom) yields high-dimensional patch descriptors for downstream analysis. Sampling~(right) synthesizes images by traversing the learned reverse SiT trajectory.}
    \label{fig:reveal}
\end{figure}

As illustrated in Fig.~\ref{fig:reveal}, REVEAL is a latent generative model that synthesizes high-fidelity medical images by aligning the feature space of a Scalable Interpolant Transformer~(SiT)~\cite{ma2024sit} with a specialized, pretrained vision encoder. The architecture operates within a compressed latent space, where an input image $x$ is mapped to $z = E(x)$ via a pretrained VAE encoder~$\emph{E}$. The generative backbone is trained to reverse a stochastic interpolation process, while a Representation Alignment~\cite{yurepresentation} strategy injects domain-specific knowledge into the student model. REPA achieves this by maximizing the cosine similarity between the pretrained teacher representation $y_*$ and the student's hidden states $h_t$, which results in

\begin{equation}
    \mathcal{L}_{\text{REPA}}(\theta, \phi) := -\mathbb{E}_{x, \epsilon, t} \left[ \frac{1}{N} \sum_{n=1}^{N} \text{sim}(y_*^{[n]}, h_\phi(h_t^{[n]})) \right],
\end{equation}

\noindent where $t$ denotes the timestep, $\epsilon$ is the added noise, $N$ corresponds to the number of patch tokens, $n$ indexes individual tokens, $\theta$ parametrizes the SiT backbone and $\phi$ the projection head. 

To maximize the transfer of anatomical detail, we adopt the iREPA~\cite{singh2025matters} strategy, which introduces two architectural refinements to the standard alignment recipe. First, we replace the conventional Multilayer Perceptron~(MLP) projection with a lightweight convolutional layer. This modification introduces a local inductive bias that preserves the spatial relationships between adjacent patch tokens, preventing the loss of high-frequency detail often observed in point-wise projections. Following observations that global components can diminish local feature contrast, we apply a spatial normalization layer~\cite{ulyanov2016instance} to the original target representations so that
\begin{equation}
    {y_*} = \frac{y - \gamma\mathbb{E}[y]}{\sqrt{\text{Var}[y] + \epsilon_s}},
\end{equation}

\noindent where $y$ is the raw teacher patch embedding, $\gamma$ is a learnable scale, and $\epsilon_s$ is a numerical stability constant. This enhancement increases the signal-to-noise ratio of local anatomical structures by sacrificing global information to improve spatial contrast, ensuring the model focuses on distinct structural signals rather than global variance.

The full training objective, combining the denoising loss with the alignment term, can be written as

\begin{equation}
    \mathcal{L} = \mathcal{L}_{\text{Denoising}} + \lambda\cdot\mathcal{L}_{\text{REPA}},
\end{equation}

\noindent where $\lambda$ is the projection coefficient that balances the two objectives. Following the iREPA implementation, $\lambda$ was set to 1. By minimizing this objective, the framework encourages SiT hidden states to match the teacher's patch-level feature geometry, effectively distilling semantic priors that support learning complex anatomical structures.

The integrated framework supports two primary operational inference-time modes, as depicted in Fig.~\ref{fig:reveal}. During training, the model optimizes joint denoising and feature alignment objectives to refine the generative trajectory. Once trained, the system enables high-fidelity sampling by traversing the learned reverse SiT trajectory to synthesize images from noise through the VAE decoder. Furthermore, the frozen SiT blocks serve as a robust feature extractor, yielding high-dimensional patch descriptors extracted from intermediate transformer layers at $t{=}0$. These features capture the aligned semantic properties of the teacher, making them suitable for downstream clinical tasks (\eg, classification). The implementation details are provided in Section~\ref{subsec:implementation} and in the publicly released code repository.

\section{Methodology}

\begin{figure}[!t]
    \centering
    \begin{subfigure}{\textwidth}
    \includegraphics[width=\textwidth]{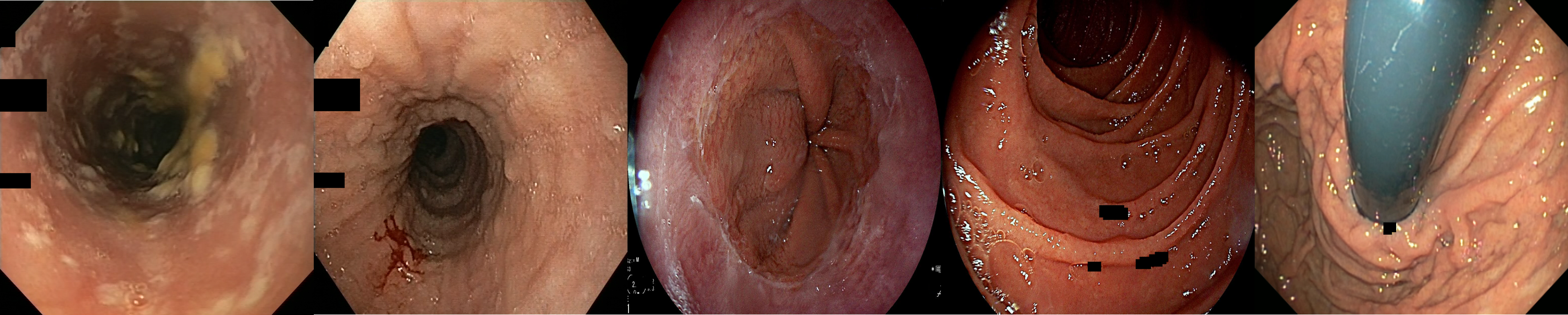}
        \caption{Samples from the GN-5M dataset.}
        \label{fig:gastronet}
    \end{subfigure}
    \vfill
    \begin{subfigure}{\textwidth}
        \centering
        \includegraphics[width=\textwidth]{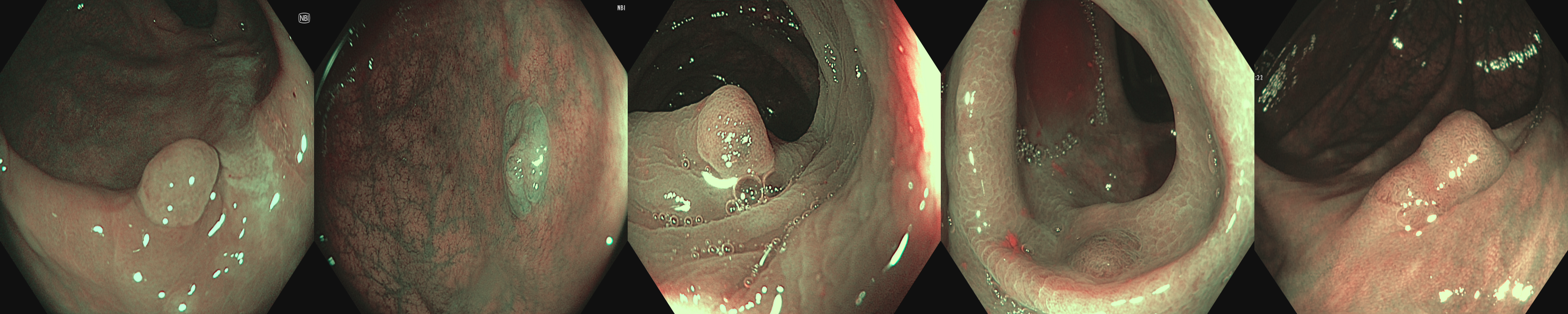}
        \caption{Samples from the POLAR dataset.}
        \label{fig:polar}
    \end{subfigure}
    \vfill
    \begin{subfigure}{\textwidth}
        \centering
        \includegraphics[width=\textwidth]{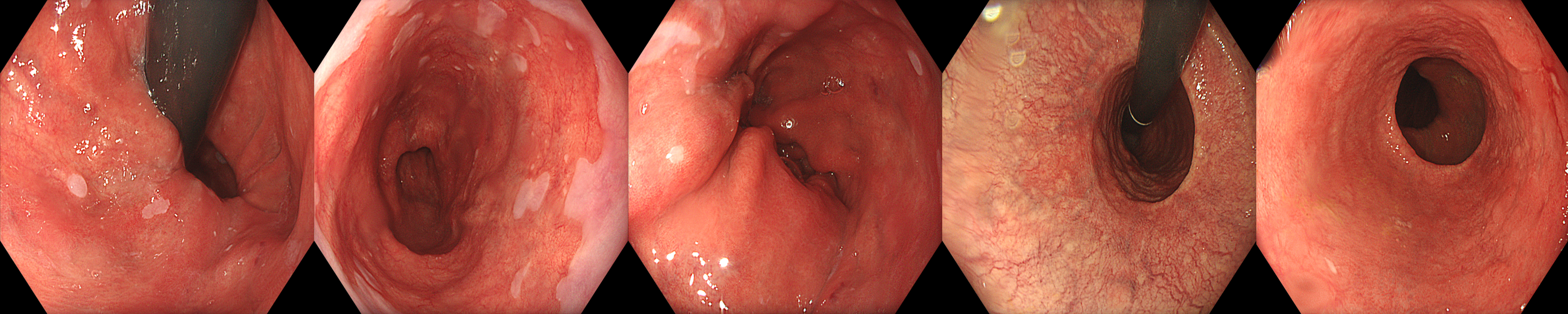}
        \caption{Samples from the BE dataset.}
        \label{fig:be}
    \end{subfigure}
    \vfill
    \begin{subfigure}{\textwidth}
    \includegraphics[width=\textwidth]{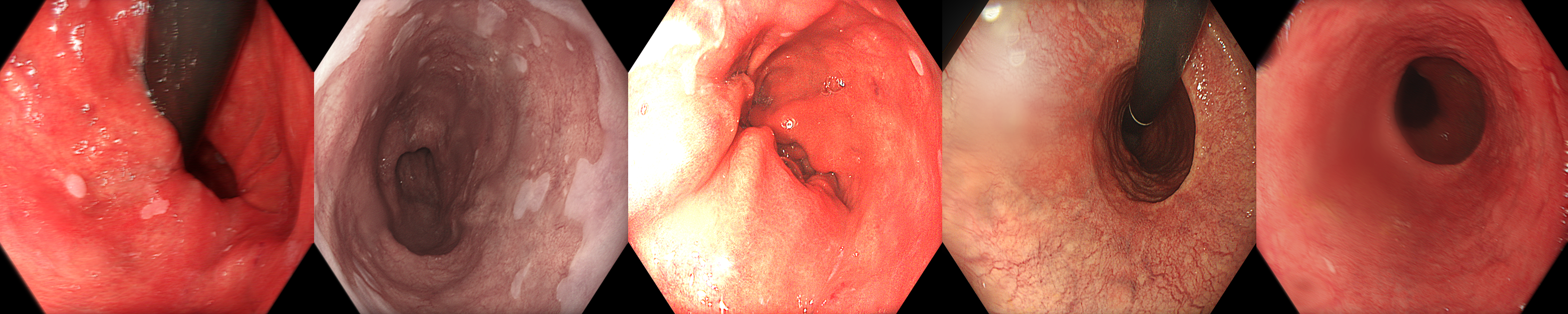}
        \caption{Samples from the corrupted BE dataset.}
        \label{fig:bec}
    \end{subfigure}
    \caption{Examples of images from the various datasets used in this study: (a)~unlabeled GN-5M images employed for training, illustrating the diversity of the data; (b)~polyp images from the POLAR dataset; (c)~sample images from the private BE dataset; (d)~examples of corrupted BE images.}
    \label{fig:datasets}
\end{figure}

\subsection{Datasets}
\label{sec:data}

\subsubsection{Pretraining Dataset.} We use the GN-5M dataset~\cite{jong2025gastronet} to pretrain our specialized vision encoders and to train our generative models. GN-5M is a large, multicenter dataset of 4,820,653 unlabeled endoscopic images collected from eight Dutch hospitals, across multiple endoscopic procedures~(colonoscopy, gastroscopy, capsule endoscopy) and imaging modalities~(white light, narrow-band, blue light, and linked color imaging), with representative samples illustrated in Fig.~\ref{fig:gastronet}. For component analysis, we rely on a representative subset of roughly 250,000 images, and for training the final high-capacity models, we use the full set.

\subsubsection{Classification Datasets.} To assess the semantic quality of the learned features, we use the publicly available POLyp Artificial Recognition~(POLAR) benchmark~\cite{houwen2023computer}, with sample images shown in Fig.~\ref{fig:polar}. The dataset is divided into a neoplastic group~(NEO), containing adenomas and sessile serrated lesions, and a non-neoplastic group~(non-NEO), consisting of hyperplastic polyps. The combined training set comprises 511 NEO patients~(1,109 polyps; 2,194 images) and 166 non-NEO patients~(230 polyps; 443 images); the test set contains 206 NEO patients~(489 images) and 67 non-NEO patients~(99 images).

Performance is further assessed on a private Barrett’s Esophagus neoplasia~(BE) dataset. The combined training set comprises 273 NEO patients~(696 images) and 44 non-NEO patients~(457 images), and the test set includes 35 NEO patients~(102 images) and 43 non-NEO patients~(171 images). The test set is enriched with subtle cases of early BE neoplasia, providing a more challenging and clinically relevant evaluation of model performance. Some samples of this dataset are available in Fig.~\ref{fig:be}.

\subsubsection{Robustness Dataset.} To assess model robustness under realistic imaging artifacts, we additionally construct a corrupted variant of the BE test set, BE-C, by applying synthetically introduced perturbations to each test image. We consider eight corruption types commonly encountered in clinical endoscopy practice: motion blur, defocus blur, 
overexposure, hue shift, saturation, contrast, sharpness, and brightness distortion. Following previous work~\cite{jaspers2024robustness}, for each test image, three corrupted copies are generated, each with a randomly sampled number of simultaneous corruptions ($1$--$5$) and 
independently sampled severity levels ($1$--$5$ per corruption), yielding a total of 819 corrupted images that form a diverse and progressively challenging robustness benchmark, as demonstrated in Fig.~\ref{fig:bec}.

\subsection{Implementation Details}
\label{subsec:implementation}

All experiments were conducted on a server equipped with four NVIDIA H100 GPUs (94\,GB VRAM each), an AMD EPYC 9334 CPU, and 768\,GB of system memory. All models were trained with mixed precision using \texttt{bfloat16}.

\subsubsection{Vision Encoders.} We employ several pretrained vision encoders as feature backbones, namely SAM2~\cite{ravi2024sam}, DINOv2~\cite{oquab2023dinov2}, DINOv3~\cite{simeoni2025dinov3}, and a DINOv2 checkpoint pretrained on GN-5M~\cite{jong2025gastronet}. To complete this set, we trained an additional DINOv3 ViT-B/16 on GN-5M via a two-stage curriculum. In the first stage, the model is initialized from a DINOv3 checkpoint pretrained on LVD-1689M~\cite{simeoni2025dinov3} and continued for 115,000 iterations using AdamW with an initial learning rate of $1\times10^{-4}$, cosine decay to $1\times10^{-5}$, a 4-epoch linear warmup, and weight decay annealed from $0.02$ to $0.1$. Augmentation is kept conservative, with 6 local crops and an iBOT masking ratio of $[0.1, 0.4]$ at a mask probability of $0.3$, to promote stable adaptation to the endoscopic domain. In the second stage, training resumes from the first stage checkpoint for a further 115,000 iterations at a reduced learning rate of $5\times10^{-5}$ without warmup, while the number of DINO and iBOT prototypes is increased to $16{,}384$, local crops to 8, and the masking ratio extended to $[0.1, 0.5]$, encouraging richer and more diverse gastroscopy-specific representations. Both stages employ RoPE positional embeddings, Sinkhorn--Knopp centering, an effective batch size of $832$ across 4 GPUs, and normalization statistics computed from GN-5M.

\subsubsection{Generative Backbone.} For the generative model, we use a base learning rate of $2\times10^{-4}$ with a linear warmup over $18{,}000$ iterations, followed by cosine decay to a final learning rate of $5\times10^{-5}$, an effective batch size of $1{,}024$, and an EMA decay of $0.9996$. The iREPA alignment module uses a convolutional projection layer with kernel size $3$. Following previous work~\cite{ma2024sit,li2025back}, we adopt a logit-normal distribution over the noise level $t$ during training: $\text{logit}(t) \sim \mathcal{N}(\mu, \sigma^2)$. Concretely, we sample $s \sim \mathcal{N}(\mu, \sigma^2)$ and set $t = \text{sigmoid}(s)$, with $\mu = \sigma = 0.8$. All images are processed at a resolution of 256~$\times$~256 pixels. The remaining hyperparameters and configurations are provided alongside the publicly released code repository.

\subsection{Experiments}

\subsubsection{Component Analysis.} The study on architectural choices is structured into four primary axes: (i)~selection of the latent space through various VAE backends, namely Stable Diffusion 2~(SD2)~\cite{rombach2022high}, Stable Diffusion 3~(SD3)~\cite{esser2024scaling}, FLUX.1-dev~\cite{flux1}, FLUX.2-dev~\cite{flux2}, and JiT~\cite{li2025back} as a pixel-space alternative; (ii)~evaluation of different target representations for alignment; (iii)~architectural scaling from SiT-S to SiT-L; and (iv)~the effect of dataset size and training duration.

Synthesis quality is quantified using FID, while representation quality is assessed through linear probing on the BE and POLAR benchmarks. For classification, features are extracted from the 8th layer of the SiT backbone at $t=0$, with results reported as the mean accuracy across five-fold cross-validation. FID scores are computed from samples generated using the Euler solver with 50 function evaluations~(NFEs).

\subsubsection{Classification Performance.} To assess the semantic richness of the learned representations, we apply a linear probing protocol on the POLAR and BE benchmarks. Features from the REVEAL backbone are benchmarked against a broad set of encoders, including general-purpose models SAM2~\cite{ravi2024sam}, DINOv2~\cite{oquab2023dinov2}, and DINOv3~\cite{simeoni2025dinov3}; DINO variants pretrained on GN-5M~\cite{jong2025gastronet}; and endoscopy-specific foundation models such as EndoViT~\cite{batic2024endovit} and Endo-FM~\cite{wang2023foundation}. We evaluate on the test sets of both benchmarks with five-fold cross-validation and report the mean AUC and AUPRC over all folds.

\subsubsection{Robustness Performance.} We follow the same linear probing protocol and encoder benchmarking as above, reporting the mean AUC and AUPRC over all folds of a five-fold cross-validation. However, we evaluate exclusively on BE-C, the corrupted variant of the BE test set described in Section~\ref{sec:data}.

\subsubsection{Qualitative results.} We provide a qualitative assessment of the images synthesized by REVEAL to evaluate visual fidelity and anatomical coherence. Samples are generated using the Heun solver with 50 NFEs. Beyond unconditional generation, we conduct inpainting and outpainting experiments as spatial robustness probes, assessing whether the learned distribution respects the geometric and textural constraints of the gastrointestinal manifold. Following RePaint~\cite{lugmayr2022repaint}, inpainting and outpainting are performed by resampling the known image regions from the forward process at each denoising step and harmonizing them with the generated unknown regions, requiring no architectural modifications or task-specific finetuning.

\begin{table}[!t]
    \centering
    \setlength{\tabcolsep}{3pt}
    \caption{Ablation of architectural components. We evaluate the impact of latent spaces, target representations, and scaling factors on image fidelity and linear probing accuracy on BE and POLAR. Legend: $^*$Trained on the full set of 5M images.}
    \label{tab:component}
    \begin{tabular}{ccc>{\centering\arraybackslash}p{1.0cm}>{\centering\arraybackslash}p{1.15cm}>{\centering\arraybackslash}p{1.15cm}>{\centering\arraybackslash}p{1.35cm}}
    \toprule
    \textbf{VAE} & \textbf{Target Rep.}& \textbf{Arch.} & \textbf{Iters.}  & \textbf{FID} & \textbf{BE} & \textbf{POLAR} \\
    \midrule
     \cellcolor{cyan!30}--- & ---  & JiT-B/16 & 150k & 28.10 & --- & --- \\
     \cellcolor{cyan!30}SD2 & ---  & SiT-B/2 & 150k & \underline{13.55} & 0.853 & \underline{0.646} \\
     \cellcolor{cyan!30}SD3 & ---  & SiT-B/2 & 150k & 14.22 & \underline{0.871} & \underline{0.646} \\
     \cellcolor{cyan!30}FLUX.1-dev & ---  & SiT-B/2 & 150k & 28.88 & 0.841 & 0.642 \\
     \cellcolor{cyan!30}FLUX.2-dev & ---  & SiT-B/2 & 150k & 14.46 & 0.824 & 0.638 \\
     \midrule
     SD2 & \cellcolor{green!30}SAM2  & SiT-B/2 & 150k & 10.98 & 0.855 & 0.623 \\
     SD2 & \cellcolor{green!30}DINOv2-B  & SiT-B/2 & 150k & 10.51 & \underline{0.872} & 0.638 \\
     SD2 & \cellcolor{green!30}DINOv3-B  & SiT-B/2 & 150k & 10.78 & 0.855 & 0.629 \\
     SD2 & \cellcolor{green!30}DINOv2-B~(GN-5M)  & SiT-B/2 & 150k & 10.71 & 0.860 & 0.653 \\
     SD2 & \cellcolor{green!30}DINOv3-B~(GN-5M)  & SiT-B/2 & 150k & \underline{10.18} & 0.865 & \underline{0.682} \\
     \midrule
     SD2 & DINOv3-B~(GN-5M)  & \cellcolor{red!30}SiT-S/2 & 150k & 13.07 & 0.863 & 0.682 \\
     SD2 & DINOv3-B~(GN-5M)  & \cellcolor{red!30}SiT-L/2 & 150k & \underline{9.33}  & \underline{0.881} & \underline{0.684} \\
     \midrule
     SD2 & DINOv3-B~(GN-5M)  & SiT-L/2 & \hphantom{$^*$}\cellcolor{yellow!30}150k$^*$ & 5.43 & 0.905 & 0.763\\
     SD2 & DINOv3-B~(GN-5M)  & SiT-L/2 & \hphantom{$^*$}\cellcolor{yellow!30}200k$^*$ & 5.37 & \textbf{0.906} & 0.768\\
     SD2 & DINOv3-B~(GN-5M)  & SiT-L/2 & \hphantom{$^*$}\cellcolor{yellow!30}250k$^*$ & \textbf{5.32} & 0.904 & \textbf{0.773}\\
    \bottomrule
    \end{tabular}
\end{table}

\section{Results \& Discussion}

\subsection{Component Analysis}

Table~\ref{tab:component} systematically evaluates the REVEAL architecture by disentangling the influence of latent space design, teacher representations, and model scaling on both generative and discriminative performance. These ablations establish a robust baseline for foundation endoscopic synthesis.

\noindent \colorbox{cyan!30}{\textbf{VAE Latent Space.}} The findings indicate that SD2 provides the optimal balance for endoscopic synthesis. While alternative models like SD3 and FLUX offer higher-dimensional representations, the compact 4-channel bottleneck of SD2 is more easily modulated by the SiT backbone, leading to superior image fidelity and competitive linear probing performance. This efficiency enables high-quality textural reconstruction without excessive computational overhead. Given the substantially inferior generation quality of the pixel-space JiT baseline, and the additional implementation effort required to support feature 
extraction in pixel space, we exclude it from further evaluation. 

\noindent \colorbox{green!30}{\textbf{Target Representation.}} The results demonstrate that representation alignment consistently improves generative performance over unguided methods. While general-purpose teachers provide competitive priors, the domain-specific DINOv3-B~(GN-5M) outperforms all alternatives in FID and POLAR. DINOv2-B achieves a marginally higher BE accuracy, likely attributable to its finer patch size and higher operating 
resolution; however, this advantage does not transfer to generation fidelity. We hypothesize that the resolution shift required by DINOv2-B~(GN-5M) during alignment results in this observed degraded generative performance. Consequently, the architectural compatibility and semantic richness of the DINOv3-B~(GN-5M) teacher make it the superior choice for capturing the complex endoscopic morphologies.

\noindent \colorbox{red!30}{\textbf{Architecture Scaling.}} Scaling the SiT backbone from Small to Large yields a substantial improvement in image fidelity. While representation quality also increases, the gain is more tempered because we extract features early at the 8th layer, which favors architectural efficiency. The most significant impact of increased parameter capacity is observed in the sharp reduction of FID.

\noindent \colorbox{yellow!30}{\textbf{Training Scaling.}} Transitioning from the 250k subset to the full GN-5M dataset yields immediate and drastic improvements across all metrics. Even at an equivalent number of training iterations, expansion to the full dataset significantly enhances image fidelity and linear probing results. The increased data diversity allows the model to cover the underlying clinical distribution better and learn more robust semantic features. Continued training yields further gains, though with diminishing returns beyond 200k iterations, suggesting the model approaches convergence on this distribution.

\begin{table}[!t]
    \centering
    \caption{Linear probing performance on downstream clinical benchmarks.}
    \label{tab:classification}
    \setlength{\tabcolsep}{4pt}
    \begin{tabular}{lcccccc}
        \toprule
         \multirow{2}{*}{\textbf{Model}} & \multirow{2}{*}{\textbf{Arch.}} & \multirow{2}{*}{\textbf{Resolution}} & \multicolumn{2}{c}{\textbf{BE}} & \multicolumn{2}{c}{\textbf{POLAR}} \\
         \cmidrule(lr){4-5} \cmidrule(lr){6-7}
         & & & \textbf{AUC} & \textbf{AUPRC} & \textbf{AUC} & \textbf{AUPRC} \\
         \midrule
         SAM2 & ViT-B/16 & 1024~$\times$~1024 & 0.583 & 0.463  & 0.684 & 0.913 \\
         DINOv2 & ViT-B/14 & 518~$\times$~518 & 0.772 & 0.673  & 0.661 & 0.898 \\
         DINOv3 & ViT-B/16 & 256~$\times$~256 & 0.765 & 0.695  & 0.737 & 0.926\\
         \midrule
         EndoViT & ViT-B/16 & 224~$\times$~224 & 0.629 & 0.475 & 0.632 & 0.892 \\
         Endo-FM & ViT-B/16 & 224~$\times$~224 & 0.764 & 0.689 & 0.683 & 0.908\\
         DINOv2~(GN-5M) & ViT-B/14 & 336~$\times$~336 & \underline{0.818} & \underline{0.779} & \underline{0.790} & \textbf{0.948} \\
         DINOv3~(GN-5M) & ViT-B/16 & 256~$\times$~256 & \textbf{0.835} & \textbf{0.790} & \textbf{0.808} & \underline{0.947}\\
         \midrule
         REVEAL & SiT-L/2 & 256~$\times$~256 & 0.786 & 0.728 & 0.758 & 0.935 \\
         \bottomrule
    \end{tabular}
\end{table}

\subsection{Classification Performance}
The semantic utility of REVEAL is evaluated through linear probing on the POLAR and BE benchmarks, as detailed in Table~\ref{tab:classification}. Baseline models lacking domain-specific features demonstrate significantly lower performance, confirming that out-of-domain priors are insufficient for resolving subtle mucosal transitions. DINOv3~(GN-5M) emerges as the strongest discriminative encoder overall, likely benefiting from its domain-specific pretraining combined with its expressive self-supervised objective. Notably, EndoViT underperforms general-purpose encoders DINOv2 and DINOv3 on both benchmarks, suggesting that domain specificity alone is insufficient 
without adequate pretraining scale and objective expressiveness.
Remarkably, despite being a generative model not explicitly optimized for discriminative tasks, REVEAL achieves these results using only the first 8 layers of the transformer backbone. Operating on a latent space compressed by a frozen, general-purpose SD2 VAE, which inherently limits the preservation of high-frequency textural detail, REVEAL still convincingly outperforms specialized models like EndoViT and Endo-FM, and surpasses all general-purpose encoders on both benchmarks, despite the fundamental differences in training objective and architecture.

\subsection{Robustness Performance}
Under corrupted imaging conditions, REVEAL demonstrates strong robustness relative to its clean-image performance, maintaining competitive representations despite the added perturbations. REVEAL once again outperforms all general-purpose encoders, including SAM2, DINOv2, and DINOv3. It further surpasses the endoscopy-specific models EndoViT and Endo-FM by a substantial margin, confirming that its representations encode clinically meaningful structure. Remarkably, EndoViT degrades catastrophically under corruption, suggesting that its representations are highly sensitive to low-level imaging artifacts despite being domain-specific. Under corrupted conditions, DINOv3~(GN-5M) again achieves the highest overall performance, with DINOv2~(GN-5M) remaining closely competitive, underscoring the robustness advantage conferred by domain-adapted pretraining. We note that REVEAL operates on features extracted from a frozen SD2 VAE encoder, a general-purpose model not designed for robustness to low-level imaging artifacts, which may introduce additional sensitivity to certain corruption types. Furthermore, REVEAL is evaluated at the standard noise level $t=0$; intermediate diffusion timesteps, which naturally denoise corrupted inputs, could be expected to yield further gains and represent a promising direction for improving robustness without any retraining.

\begin{table}[!t]
    \centering
    \caption{Linear probing performance on downstream clinical robustness benchmarks.}
    \label{tab:robustness}
    \setlength{\tabcolsep}{10pt}
    \begin{tabular}{lcccc}
        \toprule
         \multirow{2}{*}{\textbf{Model}} & \multirow{2}{*}{\textbf{Architecture}} & \multirow{2}{*}{\textbf{Resolution}} & \multicolumn{2}{c}{\textbf{BE-C}} \\
         \cmidrule(lr){4-5}
         & & & \textbf{AUC} & \textbf{AUPRC}\\
         \midrule
         SAM2 & ViT-B/16 & 1024~$\times$~1024 & 0.524 & 0.385 \\
         DINOv2 & ViT-B/14 & 518~$\times$~518 & 0.724 & 0.609 \\
         DINOv3 & ViT-B/16 & 256~$\times$~256 & 0.744 & 0.663 \\
         \midrule
         EndoViT & ViT-B/16 & 224~$\times$~224 & 0.524 & 0.400 \\
         Endo-FM & ViT-B/16 & 224~$\times$~224 & 0.692 & 0.588\\
         DINOv2~(GN-5M) & ViT-B/14 & 336~$\times$~336 & \underline{0.810} & \textbf{0.774} \\
         DINOv3~(GN-5M) & ViT-B/16 & 256~$\times$~256 & \textbf{0.814} & \underline{0.771}\\
         \midrule
         REVEAL & SiT-L/2 & 256~$\times$~256 & 0.754 & 0.679\\
         \bottomrule
    \end{tabular}
\end{table}

\begin{figure}[!t]
    \centering
    \begin{subfigure}{\textwidth}
    \includegraphics[width=\textwidth]{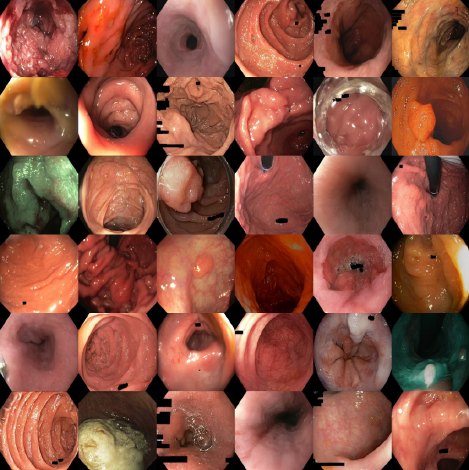}
        \caption{Unconditional samples exhibit realistic mucosal texture and
        specular highlights.}
        \label{fig:sampling}
    \end{subfigure}
    \begin{subfigure}{0.495\textwidth}
        \centering
        \includegraphics[width=\textwidth]{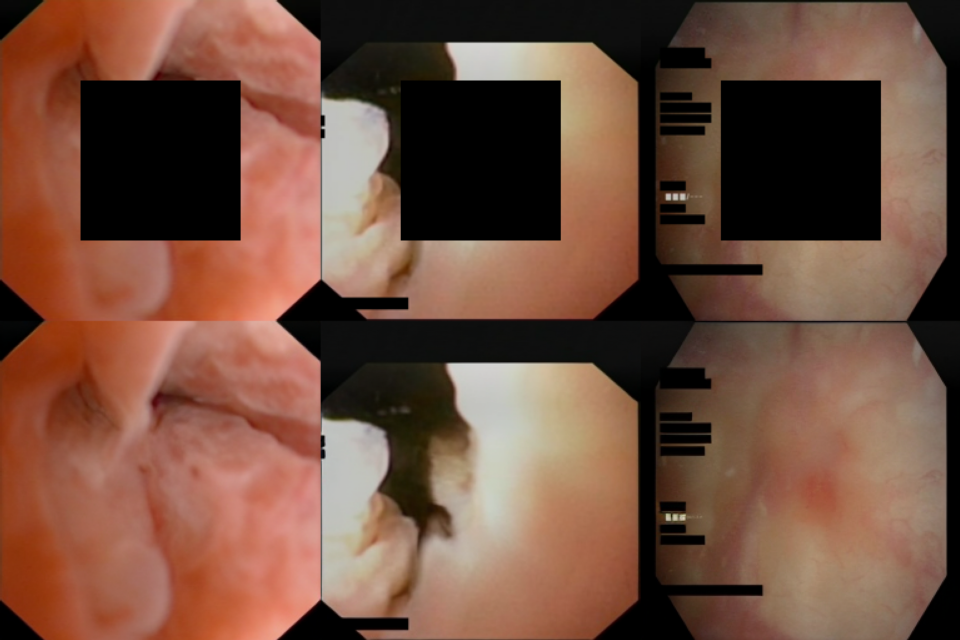}
        \caption{Inpainting reconstructs masked regions.}
        \label{fig:inpainting}
    \end{subfigure}
    \hfill
    \begin{subfigure}{0.495\textwidth}
        \centering
        \includegraphics[width=\textwidth]{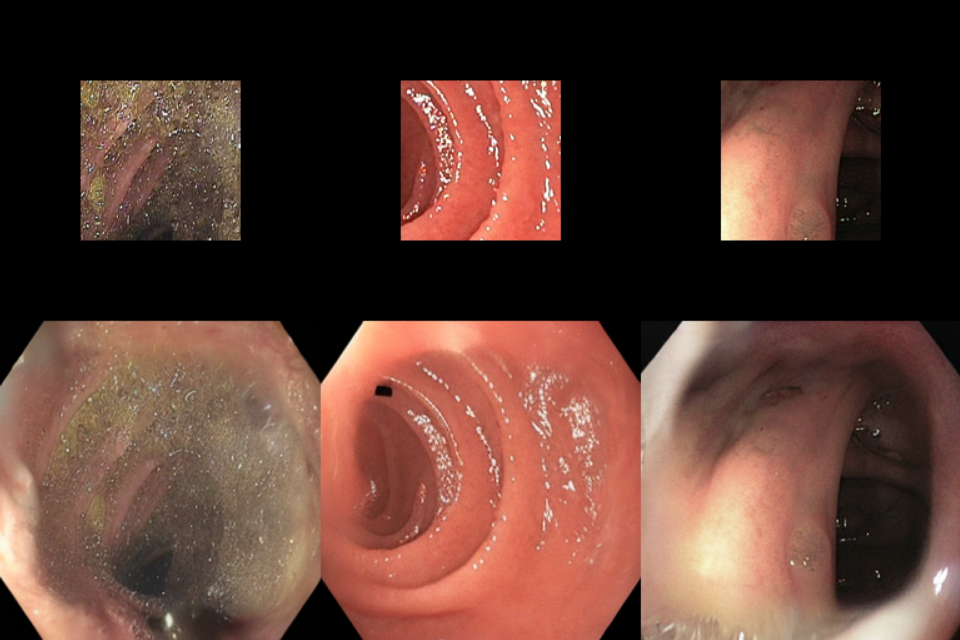}
        \caption{Outpainting extends the visible field of view.}
        \label{fig:outpainting}
    \end{subfigure}
    \caption{Qualitative results from REVEAL, generated with a 
    Heun solver using 50 NFEs. The samples demonstrate high-fidelity synthesis of diverse endoscopic scenes.} 
    \label{fig:qualitative}
\end{figure}

\subsection{Qualitative Analysis}

Figure~\ref{fig:qualitative} presents a visual assessment of images generated by REVEAL~(SiT-L/2), focusing on anatomical consistency and textural realism. The samples are produced with a Heun solver using 50 NFEs. Unconditional generation reveals that the model can synthesize plausible endoscopic scenes, capturing the intricate light reflections and mucosal textures typical of the GN-5M distribution, with diversity across anatomical regions, lesion morphologies, and imaging conditions reflecting the breadth of the pretraining data shown in Fig.~\ref{fig:gastronet}. To assess spatial robustness, we conduct inpainting and outpainting experiments. Rather than serving as clinical benchmarks, these tasks probe whether the learned distribution respects the geometric and textural constraints of the gastrointestinal manifold. REVEAL accurately reconstructs missing anatomical regions and enlarges the visible area while preserving structural continuity, indicating that aligning latent representations with a specialized teacher effectively models gastrointestinal morphology beyond the pixel level. While effective, our resampling-based strategy is known to occasionally produce harmonization artifacts at mask boundaries; more sophisticated training-free approaches, such as gradient-guided~\cite{grechka2024gradpaint} or Langevin-corrected \cite{zheng2025lanpaint} sampling, could further improve boundary consistency for inpainting and outpainting without sacrificing the simplicity of our current pipeline.

\section{Scaling \& Future Work}

The alignment of generative features with specialized clinical priors opens several avenues for further exploration. REVEAL serves as a foundation generative model in endoscopy, trained on a massive multicenter scale that is typically inaccessible to most research groups. Future iterations can leverage the pretrained weights to expand this architecture beyond its current state. 

\subsubsection{Scaling.} The current architecture can be scaled along two natural axes: resolution and model capacity. As the SD2 VAE natively supports higher-resolution inputs, scaling to finer spatial detail is a straightforward extension that would better preserve high-frequency mucosal textures. Larger transformer backbones, analogous to the scaling laws observed in natural image generation, are equally promising, as the results in Table~\ref{tab:component} demonstrate; at greater model capacity, the alternative VAE architectures explored in this work may also prove more suitable, and domain finetuning of any of them could yield additional gains. The publicly released weights provide a strong initialization point for both directions.

\subsubsection{Future Work.} The primary value of high-fidelity unconditional pretraining at this scale lies in its utility as a generative backbone for a broad set of clinically relevant downstream applications. Concretely, we envision finetuning on labeled subsets for conditional synthesis, enabling targeted pathology generation, rare lesion augmentation, and segmentation via generative features or Symmetrical Flow Matching~\cite{Caetano_2026}. The aligned latent space further offers a pathway toward generative classifiers~\cite{favero2025conditional,caetano2025medsymmflow} for diagnostic tasks, and coupling the backbone with text encoders or semantic masks enables precise conditional synthesis to address the long-tail distribution of clinical findings~\cite{zhao2025maisi}. Beyond generation, likelihood estimation over the learned distribution enables rigorous dataset coverage analysis and out-of-distribution detection, with rare pathological cases identifiable and synthesizable through targeted feature manipulation. Robustness validation across devices and centers, and segmentation benchmarks, complete our near-term roadmap. Releasing weights publicly allows the community to pursue these directions immediately.


\section{Conclusion}
We present REVEAL, a foundation generative model for gastrointestinal endoscopy trained on nearly five million clinical images from GN-5M. Through systematic ablation, we establish that representation alignment with domain-adapted visual priors is the primary driver of synthesis fidelity: in-domain encoders consistently outperform general-purpose alternatives, and scaling to the full clinical distribution yields the largest single gain across all metrics. The resulting backbone simultaneously advances generative and discriminative performance: despite carrying no supervised signal, REVEAL surpasses dedicated endoscopy foundation models such as EndoViT and Endo-FM under both clean and corrupted imaging conditions.

These results position large-scale generative pretraining as a viable and complementary path to masked image modeling for building clinical visual representations. The aligned latent space is a natural substrate for a broad range of downstream applications, such as conditional synthesis for targeted pathology generation, rare lesion augmentation, segmentation via generative features, and out-of-distribution detection through likelihood estimation. By releasing model weights publicly, we provide the community with a strong initialization for these directions, and invite further exploration of scaling, higher-resolution training, and conditional finetuning on labeled clinical subsets.

\section*{Acknowledgements}
This work used the Dutch national e-infrastructure with the support of the SURF Cooperative using grant no. EINF-13640. This publication is part of the project “You won't find what you don't image: Exposing blind spots in endoscopic cancer screening” (project number: 19091) of the research program Talentprogramma Veni-TTW which is financed by the Dutch Research Council~(NWO).

%
%
\bibliographystyle{splncs04}
\bibliography{main}
\end{document}